\documentclass[journal]{IEEEtran}
\IEEEoverridecommandlockouts
\usepackage{cite}
\usepackage{hyperref}
\usepackage{amsmath,amssymb,amsfonts}
\usepackage{mathtools}

\usepackage{algorithmic}
\usepackage{graphicx}
\usepackage{textcomp}
\usepackage{xcolor}
\usepackage{svg}

\usepackage[]{subcaption}
\usepackage{xcolor}
\usepackage{epsfig} % for postscript graphics files
\newcommand{\Mod}[1]{\ \mathrm{mod}\ #1}    
\DeclareMathOperator*{\argmax}{arg\,max}

\def\BibTeX{{\rm B\kern-.05em{\sc i\kern-.025em b}\kern-.08em
    T\kern-.1667em\lower.7ex\hbox{E}\kern-.125emX}}
\begin{document}

\title{Feature Sharing and Integration for Cooperative Cognition and Perception with Volumetric Sensors
\thanks{This research was supported in part by the National Science Foundation under CAREER Grant 1664968 and by Toyota Motor Corporation.}

}
\author{\IEEEauthorblockN{Ehsan Emad Marvasti\IEEEauthorrefmark{1}, Arash Raftari\IEEEauthorrefmark{1}, Amir Emad Marvasti\IEEEauthorrefmark{1}, Yaser P. Fallah\IEEEauthorrefmark{1}, \\ Rui Guo\IEEEauthorrefmark{2} and Hongsheng Lu\IEEEauthorrefmark{2}}
\IEEEauthorblockA{\IEEEauthorrefmark{1}Department of Electrical and Computer Engineering, Department of Computer Science\\
University of Central Florida, Orlando, Florida, USA\\
\{e\_emad, raftari, a\_emad\}@knights.ucf.edu} yaser.fallah@ucf.edu
\IEEEauthorblockA{\IEEEauthorrefmark{2} InfoTech Labs, Toyota Motor North America, Mountain View, California, USA\\
 \{rguo, hlu\}@us.toyota-itc.com }
}

\maketitle

\begin{abstract}
% The recent advancements in communication and computational systems have led to significant improvement of situational awareness in connected and autonomous vehicles. Computationally efficient neural networks and high speed wireless vehicular networks have been some of the main contributors to this improvement. However, scalability and reliability issues caused by inherent limitations of sensory and communication systems are still challenging problems. In this paper, we aim to mitigate the effects of these limitations by introducing the concept of feature sharing for cooperative object detection (FS-COD). In our proposed approach, a better understanding of the environment is achieved by sharing partially processed data between cooperative vehicles while maintaining a balance between computation and communication load. This approach is different from current methods of map sharing, or sharing of raw data which are not scalable. The performance of the proposed approach is verified through experiments on Volony dataset. It is shown that the proposed approach has significant performance superiority over the conventional single-vehicle object detection approaches.

The recent advancement in computational and communication systems has led to the introduction of high-performing neural networks and high-speed wireless vehicular communication networks. As a result, new technologies such as cooperative perception and cognition have emerged, addressing the inherent limitations of sensory devices by providing solutions for the detection of partially occluded targets and expanding the sensing range. However, designing a reliable cooperative cognition or perception system requires addressing the challenges caused by limited network resources and discrepancies between the data shared by different sources. In this paper, we examine the requirements, limitations, and performance of different cooperative perception techniques, and present an in-depth analysis of the notion of Deep Feature Sharing (DFS). We explore different cooperative object detection designs and evaluate their performance in terms of average precision. We use the Volony dataset for our experimental study. The results confirm that the DFS methods are significantly less sensitive to the localization error caused by GPS noise. Furthermore, the results attest that detection gain of DFS methods caused by adding more cooperative participants in the scenes is comparable to raw information sharing technique while DFS enables flexibility in design toward satisfying communication requirements. 

\end{abstract}


 \begin{figure*}[!t]
\centering
 \includegraphics[width=.8\textwidth,trim={0mm 0mm 0mm 0mm},clip]{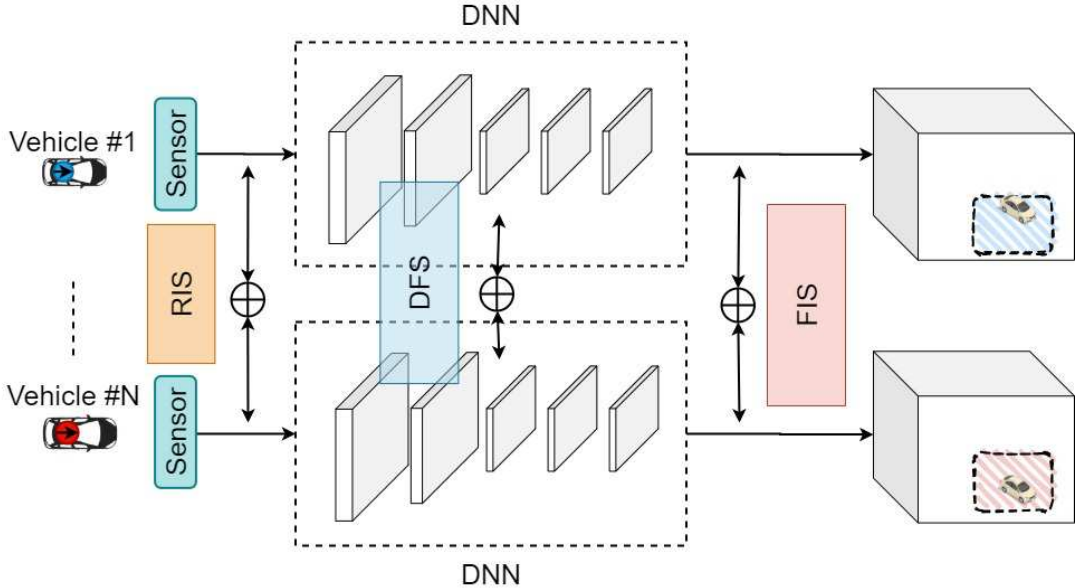}
 \caption{Overview of cooperative information aggregation methods}
 \label{fig:information aggregation approaches}
 \end{figure*}
\section{Introduction}

% The need for situational awareness has led to improvement in perception methods. However, the methods introduced are based on the existence of single observer in the environment.
% Cooperative perception in automotive applications is one of the emerging subjects leading to situation awareness performance. The incentive to study such methods arises from the 

Performance and safety of vehicle automation systems is heavily dependent on the quality of situational awareness obtained through vehicle sensors and sensor processing methods. Thus, addressing limitations of sensors and processing methods through the use of communication and cooperation between vehicles has been the subject of numerous studies in recent years. Most modern sensor processing methods rely on Deep Neural Networks (DNNs) \cite{Redmon_2016_CVPR,li2016vehicle,feng2018towards,beltran2018birdnet,10.1007/978-3-319-46448-0_2}. DNNs offer computationally efficient methods for inference systems such as object detectors, pose estimators, and motion predictors.
These DNN based inference systems can be utilized for designing effective real-time safety applications in the vehicular domain.
Since the quality of the input sensory data highly affects the performance of the DNN based applications, there were constant efforts to improve the quality of sensory devices.
Such efforts have not overcome certain limitations such as non-line of sight and range restrictions.
In addition, the use of expensive high-end sensory devices is in contrast with many low-cost manufacturing strategies in the vehicular industry.
With the introduction of high speed vehicular wireless networks, Vehicle to Everything (V2X) technologies offer the possibility of exchanging sensory information between road participants.
The vehicular wireless network is the platform for the introduction of emerging technologies such as cooperative perception contributing to the enhancement of situational awareness.
Cooperative perception enables the vehicles and infrastructure to share their acquired information leading to synergistic improvements in perceiving the environment.
As a result, the desirable level of situational awareness can be achieved by deploying sensory units at a significantly lower cost.
However, utilizing a robust cooperative perception requires efficiently addressing the challenges such as scalability issues caused by the limited bandwidth of vehicular networks and inconsistencies of shared information gathered by distributed sources.
In general, cooperative perception methods can be categorized by the type of data shared among participants.
Deployment of each category determines the bandwidth consumption and cooperative perception performance. It is desirable to choose the category where the information is efficiently transmitted with respect to the available bandwidth.
We define the categories as follows:(1) raw information sharing, (2) partially-processed information (feature) sharing (3) fully-processed information sharing. 
The raw information sharing (RIS) cooperative perception methods require more allocated bandwidth, while, in theory, yielding a better perception performance.
On the other hand, in the fully processed information sharing (FIS) method, the cost of communication is significantly lower. However, the detection of partially observed objects and resolving the lack of consensus between cooperative entities is challenging.
Therefore, it is expected to achieve lower perception performance.
The notion of Deep Feature Sharing (DFS), presented in \cite{marvasti2020bandwidthadaptive,marvasti2020cooperative}, attempts to find a middle ground between the rich information content provided by RIS and the low communication requirement of FIS.
DFS proposes a framework for extraction, transmission, and integration of partially processed data (features) obtained from intermediate layers of a neural network.
Fig.\ref{fig:information aggregation approaches} demonstrates an overview of the cooperative information aggregation approaches.
While RIS and FIS methods have been the subject of numerous studies, the cooperative techniques based on deep feature sharing have only gained attention in a couple of recent studies \cite{10.1145/3318216.3363300,marvasti2020cooperative,wang2020v2vnet,marvasti2020bandwidthadaptive}.

In this paper, we attempt to investigate the limitations, requirements, and performance of these techniques by considering a realization for each of the aforementioned cooperative perception categories and further generalizing the concept of DFS.
Some realizations of DFS approach for object detection using LIDAR are implemented and studied in this paper. These realization are constructed out of the architecture proposed in our earlier works \cite{marvasti2020cooperative, marvasti2020bandwidthadaptive} by changing some aspects or components of the original design. 
% our earlier work proposed in \cite{marvasti2020cooperative}: Feature sharing cooperative object detection (FS-COD), is considered in this paper.
Moreover, a realization of FIS approach by using LIDAR observations is presented and denoted as Hypothesis Sharing Method (HSM) in this paper.
Additionally, a LIDAR based RIS method is developed and evaluated. 
The performance of these methods has been evaluated in the presence and absence of GPS noise to measure their sensitivity to localization error.
The rest of the paper is organized as follows. In section \ref{section:Bckgroung}, a literature review on cooperative perception methods is presented.
In section \ref{section:aggregation methods}, we discuss each cooperative method along with its main building blocks while emphasizing on DFS framework and it's realizations.
In section \ref{section::experiments}, the details of the cooperative dataset along with the experimental setup are discussed. Subsequently, the analysis of the results inferred from the experiments is provided.
Finally, section \ref{Section:Concluding Remarks and Future work} concludes this work. 
\begin{figure*}[!t]
\centering
 \includegraphics[width=1\textwidth,trim={0mm 0mm 0mm 0mm},clip]{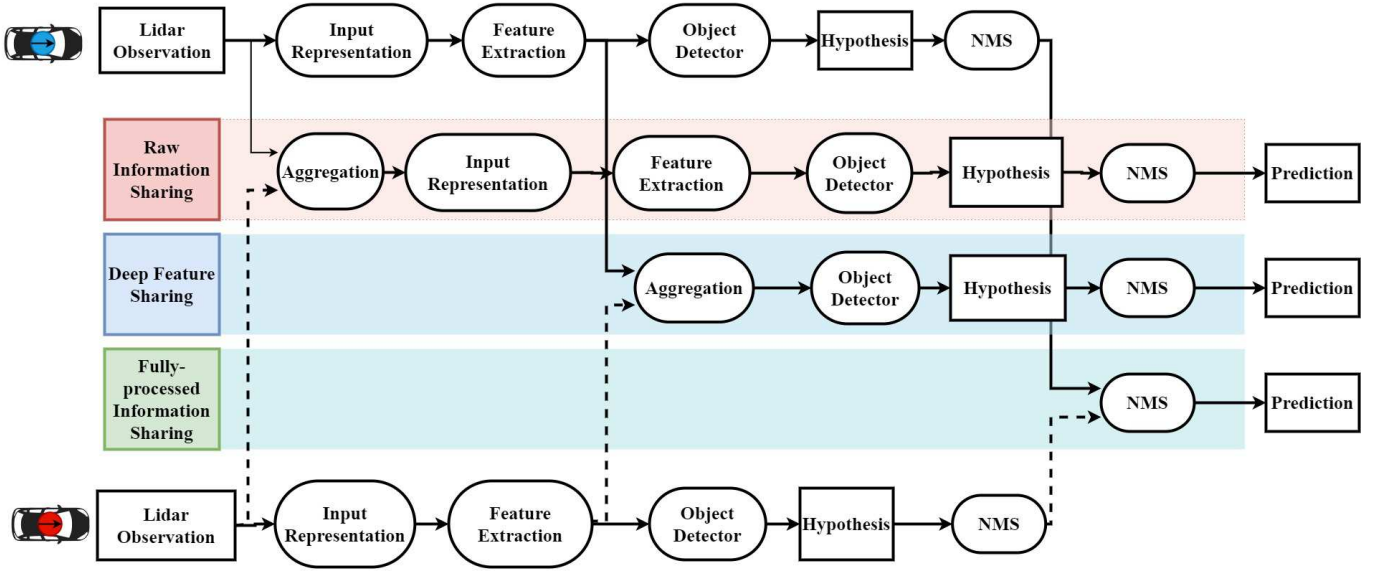}
 \caption{Overview of Lidar based cooperative object detection methods}
 \label{fig:overall}
 \end{figure*}
\section{Background}
\label{section:Bckgroung}
% \textcolor{red}{it should be written after the paper is readsy}
Cooperative perception has been the subject of many studies prior to its application in the vehicular industry through the concept of connected and autonomous vehicles (CAV). The prevalent method of sharing raw information is a common approach taken to improve situational awareness.
In \cite{6866903}, to provide a far sight satellite view to the driver, a multimodal cooperative perception method based on sharing raw sensory data is introduced. 
In \cite{10.1145/3210240.3210319}, an Augmented vehicular reality method is proposed to broadens cooperative vehicles' visual horizons.
In \cite{9149662}, a joint computation offloading and resource allocation optimization algorithm is proposed for augmented vehicular reality systems considering computation and spectrum restrictions. This algorithm optimally divides the task of processing the raw sensory data between cooperative vehicles and infrastructure units and decides whether to share raw or processed visual data between participants by considering the restrictions on computation resources of each participant and available communication bandwidth.
In \cite{8885377}, a cooperative point-cloud object detection method is proposed. The method consists of a scheme for sharing and aggregating region of interest LIDAR data, followed by a Convolutional Neural Network (CNN) for detecting objects. 
In \cite{8403778}, the communication bandwidth requirement and benefits of sharing sensory information in terms of sensing redundancy and range have been studied. 
In \cite{8417769}, the effect of deploying sensing and communication equipment at roadside infrastructure on sensing performance has been studied.

However, the limitation in communication bandwidth is still a major obstacle to deploy these methods in delay-sensitive vehicular safety applications. this limitation was the motive for the introduction of frameworks on which more practical methods can be designed.
Concepts such as Cooperative Sensing Message (CSM)\cite{CSM}, Collective Perception message (CPM)\cite{CPM}, and Environmental Perception Message (EPM)\cite{7835930} are examples of such efforts. 
Although these efforts propose different packet structures, they have a similar purpose to provide a framework for sharing and aggregation of fully processed information in the form of object lists. 
In addition, there were some efforts from the map-sharing perspective to achieve cooperative perception using FIS methods. In \cite{8280503}, a map sharing technique combined with a content control strategy was proposed to extend 3D maps created from LIDAR observations and enhance position tracking performance. 
In \cite{6953213}, an adaptive communication strategy was proposed for FIS methods to satisfy the communication bandwidth restrictions.
In \cite{8891157}, a graph matching scheme was proposed to build a 3D representation of the environment. The method efficiently maps the graph of objects detected by cooperative vehicles and infers the location of occluded objects.
A method was proposed to enhance the object detection confidence by aggregating the object detection's hypotheses of cooperative vehicles leading to build a more accurate view of the environment in \cite{8500388}
As mentioned in the previous section, these methods cannot efficiently address the challenges caused by partial occlusion and lack of consensus between cooperative participants.
These challenges were the motive to further investigate and design practical methods with the capability of adapting to current limitations and flexibility in development as these limitations fade with the advancement in technology.
The introduction of cooperative methods based on sharing and integrating intermediary representations of DNNs is an example of such efforts.
In \cite{marvasti2020cooperative}, a cooperative object detection method, FS-COD, is proposed. In this method, cooperative nodes, equipped with LIDAR, exchange and integrate the features extracted from a multi-layer CNN to cooperatively detect the objects in the environment.
\cite{marvasti2020bandwidthadaptive} proposes a bandwidth adaptive cooperative object detection method along with a shared data alignment procedure to be utilized in feature sharing methods.
Similarly, in \cite{10.1145/3318216.3363300}, a LIDAR-based feature sharing cooperative object detection method is introduced and the effectiveness of sharing a selected portion of feature-map channels is investigated.
In \cite{wang2020v2vnet}, a cooperative method is proposed for joint perception and motion prediction by sharing the compressed features extracted from LIDAR observations.
In this method, a Graph Neural Network (GNN) is trained to align and aggregate the information received from different cooperative entities at different points in time and with different viewpoints.
In general, the aforementioned feature sharing cooperative perception methods are mainly distinguished by their shared data alignment procedures and aggregation method. 
In this paper, the effect of some of these aggregation and alignment methods on the performance of cooperative object detection is studied.
\begin{figure*}[!t]
\centering
\includegraphics[width=.98\textwidth,trim={0mm 0mm 0mm 0mm},clip]{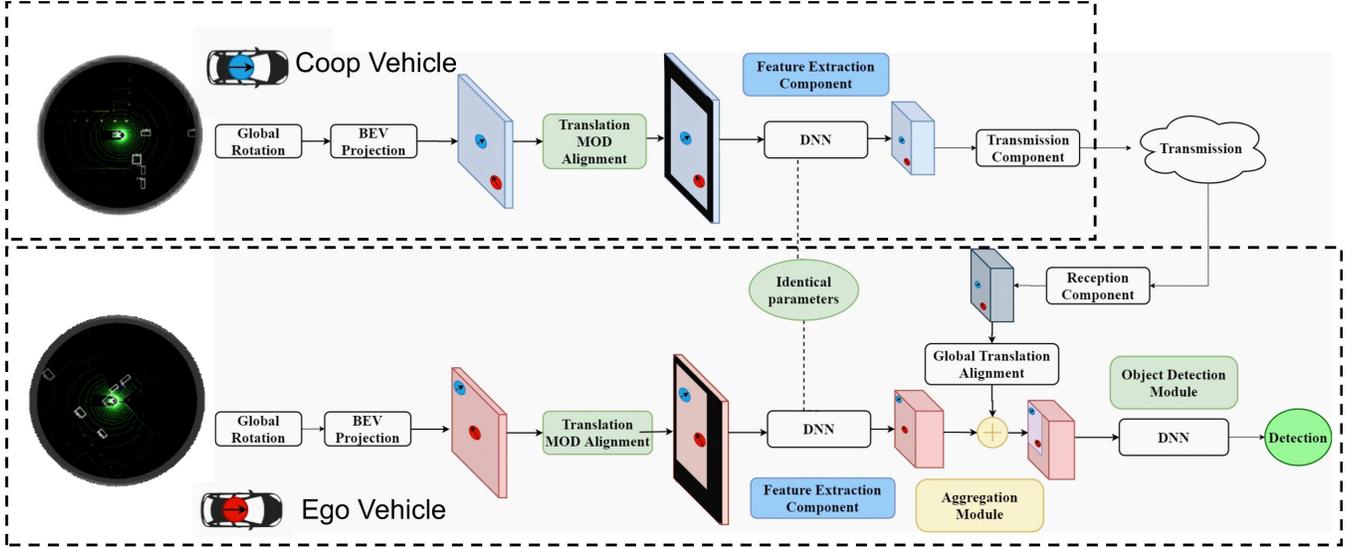}
\caption{The overview of the DFS approach for cooperative object detection using LIDAR.}
\label{fig:FSCOD-arch}
\end{figure*}
\section{Information Aggregation In Cooperative Methods}
\label{section:aggregation methods}
As discussed in the previous section, the cooperative perception methods can be distinguished by the type of data shared between participants and therefore categorized to Raw, Deep Feature, and Fully-processed Information Sharing.
In the RIS methods, the information shared between entities belongs to the sensory vector space. On the other hand, in the DFS methods, an intermediate state space is defined to embed the sensory information and share it among participants. In the FIS methods, information is embedded in the hypothesis state space defined by the task and then shared among cooperative participants. These hypotheses represent semantic attributions that are interpretable.
Therefore, many fusion algorithms and cooperative vehicular applications have been introduced using the FIS approach\cite{6232130, 6758572, 6728345}. However, the hypothesis space in the FIS approach may not necessarily be optimal with regards to fusion and compression in V2V joint perception applications. This is due to the fact that hypothesis space is defined solely for the task of detection and does not necessarily include the information needed to optimally perform the fusion. 
In other words, such state space is not defined with respect to the task of compression or fusion and therefore the method may suffer from redundancies or information loss.
On the other hand, intermediate state space can be tuned to increase the performance of compression, fusion, and any other measurement crucial for a defined task.
However, the intermediate state space is not interpretable, and therefore performing operations such as alignment on the feature vector is not trivial.
On the contrary, in the RIS methods, the shared information can suffer from heavy redundancies, while, in many cases, such redundancies are not beneficial for the defined task. As an example, the information captured from the shapes of surrounding buildings may not be useful for the detection of pedestrians in a scene. 
It is worth mentioning that in some cases, redundant information may be helpful to boost performance. For example, redundant information can be used to increase the performance of object detection in scenarios where the data gathered by sensory devices in the environment is heavily affected by noise.
However, even in scenarios where redundant information is helpful, the cost of allocating the required communication bandwidth to transfer such information can be higher than the gain in gradually increasing the performance.

The discussion further motivates us to provide a relatively comprehensive study on the performance of each category of cooperative object detection. Additionally, cooperative methods can be employed for other tasks such as motion prediction. In the rest of this section, we first introduce the components that are concertedly present in all cooperative perception approaches, followed by the proposed methodologies to develop a cooperative object detection method for each approach. Fig.\ref{fig:overall} illustrates an overview of LIDAR based cooperative object detection approaches along with their essential components. 

It's worth mentioning that in all proposed realizations, we assume each cooperative vehicle is equipped with an on-board volumetric sensory device, e.g. LIDAR, and it has access to its location and heading using a GPS device.
\begin{figure*}[!t]%
\centering
\subcaptionbox[]{\label{fig:MA}}{%
\includegraphics[width=0.40\textwidth,trim={0mm 0mm 0mm 0mm},clip]{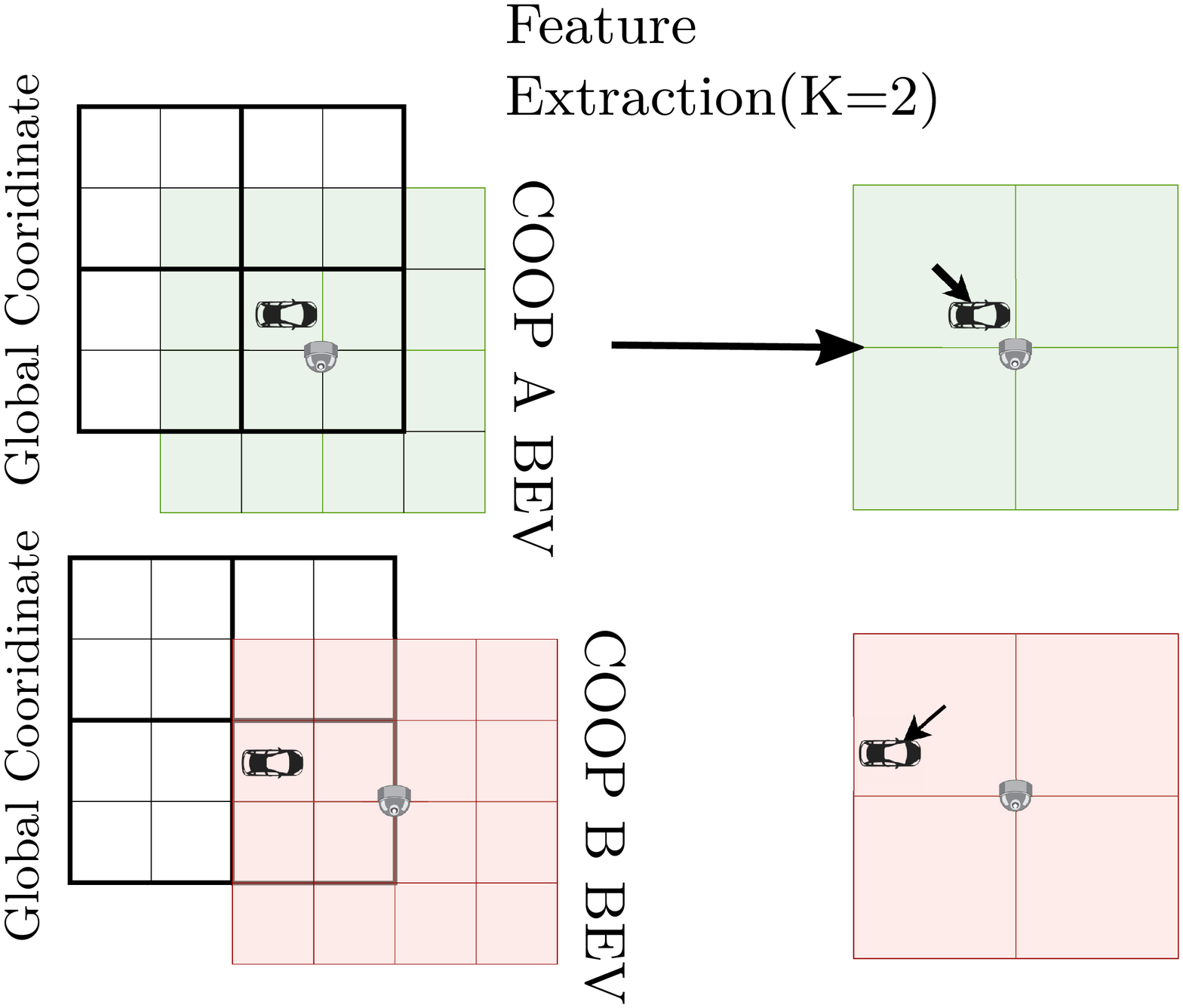}}%
\subcaptionbox[]{\label{fig:TMA}}{%
\includegraphics[width=0.58\textwidth,trim={5mm 25mm 10mm 15mm},clip]{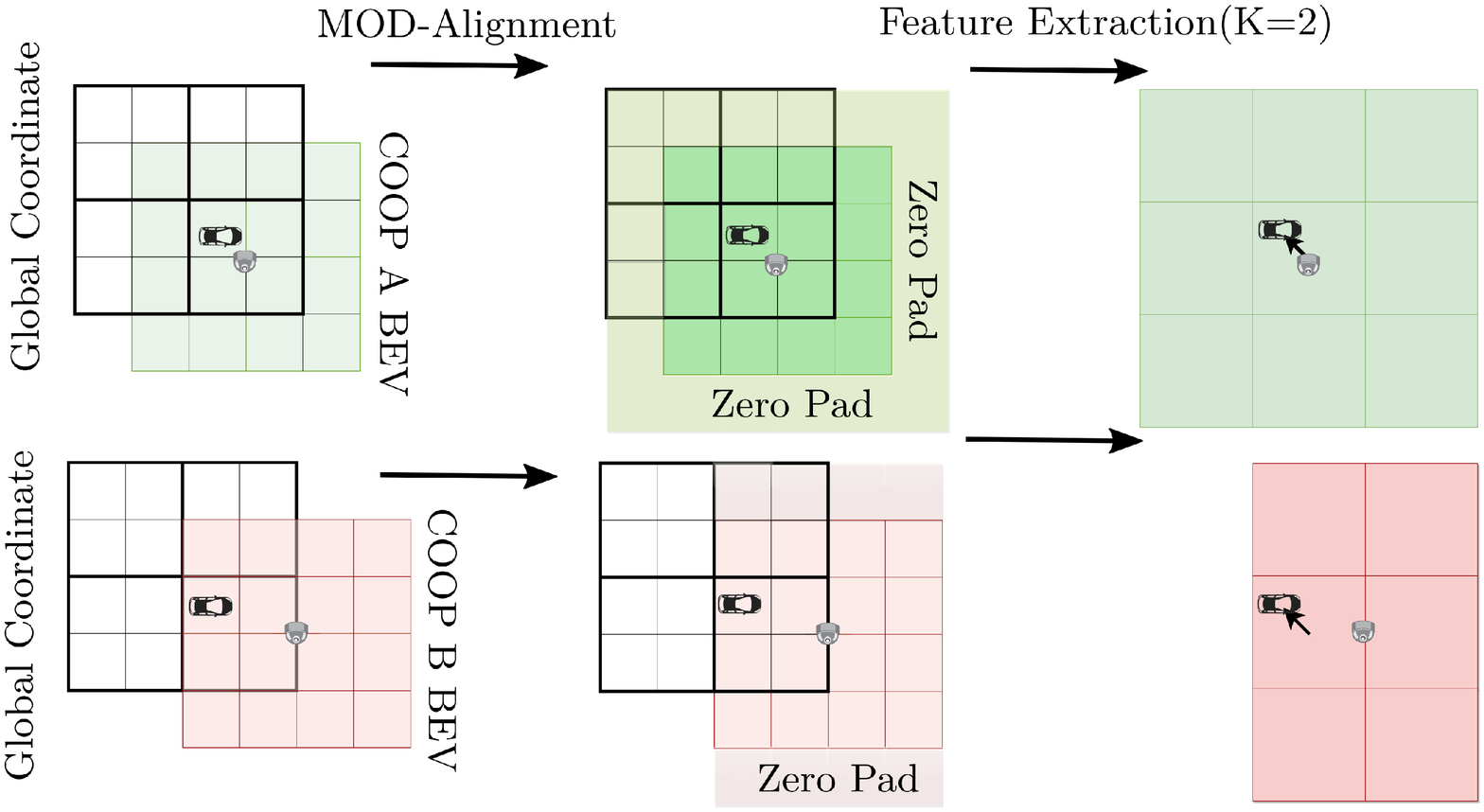}}%
\caption{ (a) An illustration of how down-sampling cause alignment error since the optimum localization vectors resulted from feeding the each fixel to CNN are contradictory. Hence, fusion of the fixels would result in localization error. K is the down-sampling rate (b) An illustration of Translation MOD-Alignment procedure to eliminates the contradiction caused by down-sampling. By padding the input images, corresponding fixels represent identical areas in a global coordinate system.}
\label{fig:TMA-NMA}
\end{figure*}
\subsubsection{Object Detection Module}
Object Detection Module (ODM) is a deep neural network (DNN),  translating the information from the sensory or intermediate feature spaces to the defined interpretable hypothesis space.
The hypothesis structure is imposed by the loss function used to train the parameters of the object detection module. In single-shot networks,  the hypothesis has the generic form of vector H, defined as follows:
\begin{equation}
        \hat{H} = [\delta \hat{x},\delta \hat{h}, \hat{\omega}, \hat{p_c}].
\end{equation}
Where $\delta x$, $\delta h$, $\omega$ and $P_c$ are representing localization, shape, orientation and classification prediction. Consequently, the output of the ODM is a tensor where each pixel (voxel) represents a hypothesis with the aforementioned variables as the prediction.
The $\delta$ notation is used where the predicted parameters are the distances of the object characteristics from predefined parameter values, e.g., predicted width of an object is $\delta \hat{w}+W$. Where $W$ is a predefined value.
% The question of necessity of using an anchor as defined in yolo is further discussed in next sections.
The loss function commonly used for estimation of such parameters is defined as:
\begin{equation}
    L = \sum_i \lambda_i D(\hat{P}_i,P_i)
\end{equation}
where $D$ is the distance measure, $\lambda_i$ is the hyper-parameters defining the importance of each term and $P$ represents the target vector.
In general, $l_2$ norm and cross-entropy are considered as the distance measure for regression and classification problems respectively.
In single-shot object detection methods, the ODM network is not designed to merge the overlapping hypotheses. Thus, the network may suggest multiple hypotheses with regards to a single object in the scene. To resolve this issue, a method, called non-maximum suppression, is required to merge overlapping detection hypotheses. The non-maximum suppression, which is mostly heuristic, can be considered as an aggregation method discussed in the next section.   
% In single-shot object detectors, the ODM does not have the capability to merge overlapping hypotheses and therefore multiple hypotheses can predict a single object in the environment.
% In order to resolve such issue the non-maximum suppression method is used to merge multiple detection's hypotheses.
% We identify non-maximum suppression component as an aggregation component and discuss the method in the next section.
% In Yolo and SSD only the pixels where the center of the object resides are set to be an object pixels. Therefore, the neighboring pixels regardless of observing the object are suppressed and forced to predict a non-object class in training.

% Additionally in YOLO, $\delta x$ is the output of sigmoid function and therefore the localization vector produced by corresponding pixel cannot predict the location of center of an object residing in neighboring pixels. 

\subsubsection{Alignment and Aggregation Module}
\label{Section:overallaggregation}
Regardless of the category of shared data, in any cooperative method, a procedure is required to align and aggregate the information gathered by spatially distributed participants from the scene. In other words, The alignment and aggregation module is the component responsible to fuse the information gathered from different receptive fields in the defined state space. A proper choice of this procedure can significantly influence the performance of the system.
while the alignment procedure in RIS and FIS methods is as simple as translation and rotation transformations, in the DFS methods, the alignment of shared features requires more effort. The Translation MOD-alignment procedure, introduced in \cite{marvasti2020bandwidthadaptive}, is an attempt to enable accurate integration of shared feature grids by eliminating inherent localization error caused by down-sampling in the neural networks.

In addition, While the aggregation procedure in the RIS methods is a trivial superposition function, the cooperative perception systems based on FIS and DFS methods are mainly differentiated by the techniques used for aggregation of shared data.
% The aggregation Module is the component responsible to fusion of information gathered from different receptive fields in the defined state space.
% The methodologies of information aggregation differs based on the category of information sharing.
The aggregating function of the cooperative methods' realizations, presented and evaluated in this paper, are discussed in the following sections.

\subsubsection{Sensory to Input Representation Component}
\label{section::Sensory to input rep}
The representation of information gathered from sensory systems determines the computational efficiency and performance of the inference systems; however, there is a trade-off between high computational efficiency and high task performance, e.g. object detection performance. Therefore, the input representation should be chosen by considering the computation limitations, task objectives, and the nature of the application. 
In our setup where the sensory data are point-clouds, gathered by LIDAR devices, projecting the information onto 3D voxel tensors prevents loss of information with regards to the task of detection. However, processing the 3D tensors requires the use of computationally costly 3D CNNs.
In the vehicular domain, since the objects of interest, i.e. vehicles and pedestrians, lie on a planar surface with a known height range, bird-eye view (BEV) projection of point-clouds is a common approach to reduce the computation complexity while not significantly affecting the performance of object detection\cite{kim2019enhanced,yang2018pixor,simony2018complex, 8088147}. Moreover, in BEV projected images, the size of the observed object remains constant and does not change with respect to its distance from the observer. This merit further makes BEV images projected from LIDAR point-clouds an appropriate choice for input data representation in cooperative methods where data gathered with different viewpoints are shared among participants.

\subsection{Raw Information Sharing Methods}

In RIS methods, the shared information is either raw data in the sensory state space or the projected raw data by using a transformation. The transformation can be a handcrafted function such as BEV projection or a neural network compressing the raw information such as an autoencoder. The cooperative methods based on the transmission of compressed data using autoencoders is similar to feature sharing methods; however, the compression methods, whether using DNNs or handcrafted functions, are not optimized to provide the maximum relevant information regarding the task, e.g. object detection. Therefore, even though the shared data can be the output of DNNs, we categorized such methods as RIS. 

In general, RIS methods require higher communication bandwidth compared to the other approaches resulting from the extensive redundancies in the transmission of raw information.
Additionally, part of the information shared between participants is irrelevant to the designated task, and therefore sharing this information does not necessarily lead to an improvement in the task's performance. 
Moreover, in the case of using a handcrafted transformation function, the process of designing an input state space capable of efficiently embed the information requires many trials.
In this paper, to compare the performance of sharing raw information to other approaches, an RIS method is designed based on sharing point-clouds between cooperative vehicles. However, the aggregated point-cloud is projected to BEV images to be used as input of ODM due to the merits of BEV projection discussed in the previous section.

\subsection{Fully-Processed Information Sharing Methods}
In general, in FIS methods, each cooperative vehicle creates a list of hypotheses based on its own observations from the environment and share it with other participants. This list usually contains information regarding the detected objects such as the classes of the detected objects, localization information of the objects, and detection confidences along with meta-data of the observer, e.g. observer's position and heading, type and location of the sensors. At each cooperative vehicle, these lists are aggregated using a fusion algorithm and the decisions are made based on the collected perception. Therefore, the FIS methods mainly differ in their fusion algorithm and the structure of the hypotheses shared between participants.

Theoretically, the FIS methods require significantly lower communication bandwidth compared to other approaches. However, these methods deliver relatively lower task performance and are mainly suitable in scenarios where available bandwidth for each cooperative vehicle is strictly limited.

In this paper, we developed an FIS method, denoted as Hypothesis Sharing Method (HSM), by borrowing the hypothesis structure of single-shot object detector methods. In addition, the greedy Non-Maximum Suppression (NMS) method utilized in \cite{Redmon_2016_CVPR}, is considered as the fusion algorithm to aggregate the overlapping hypotheses.
Although the NMS method is commonly used to merge the overlapping hypotheses gathered from an input image, the same method can be utilized to aggregate hypotheses gathered from multiple observations, e.g. multiple vehicles.
It's worth mentioning that greedy NMS methods are mainly heuristic and the filtering parameters are not optimized during training. \cite{hosang2017learning} provides a solution to perform NMS with tunable parameters. The proposed method can also be utilized as a fusion algorithm in FIS methods.

To further enhance the performance of FIS methods, both the hypothesis structure and NMS method can be subjected to further investigation and improvement.

\subsection{Deep Feature Sharing Methods}
Sharing the intermediary representations of DNNs is an alternative method for achieving cooperative perception.
Contrary to FIS and RIS, in DFS methods, the feature-space is optimized to efficiently embed and fuse information with respect to the task through training procedure.
In general, any DFS method consists of the following modules: sensory to input representation module, feature extraction component, transmission/reception component, alignment and aggregation component, and object detection module. 
Fig.\ref{fig:FSCOD-arch} provides an overview of our proposed realization of the DFS approach, along with its main components.
In the rest of this section, we will elaborate on these modules, except for sensory to input representation and object detection modules which are discussed in section \ref{section::Sensory to input rep}. 

% \begin{figure}[!pt]%
% \centering
% %
% \subcaptionbox[]{}{%
% \includegraphics[width=0.1559\textwidth]{Results/VOLONY/VOLONYRGB.eps}}%
% \subcaptionbox[]{}{%
% \includegraphics[width=0.159\textwidth]{Results/VOLONY/VOLONYPC.eps}}%
% \label{fig:third}%
% \subcaptionbox[]{}{%
% \includegraphics[width=0.148\textwidth]{Results/VOLONY/VOLONYBEV.eps}}%

% \caption{ (a) Illustration of CARLA work-space in an arbitrary scenario (b) Point-cloud generated from LIDAR device of a cooperative vehicle (c) BEV representation of the point-cloud}
% \label{fig:foobar}
% \end{figure}

% \begin{figure*}[t!]%
%   \centering
%   \qquad
%   \subcaptionbox[]{}{%
%   \includegraphics[width=1\linewidth,trim={20mm 0mm 30mm 10mm},clip]{Results/Ablation/GPS_Noise/V2/GPS.eps}}%
% %   \subfigure{{\includegraphics[width=1\linewidth,trim={0mm -5mm 0mm 6mm},clip]{Results/40m/coop8f.eps} }}%
% \caption{The effect of GPS (location) noise on Average Precision of different cooperative object detection approaches under different IoU thresholds. }
% \label{fig::GPSnoise}%
% \end{figure*}

\begin{figure}[!pt]%
  \centering
  \includegraphics[width=.5\linewidth,trim={0mm 0mm 0mm 0mm},clip]{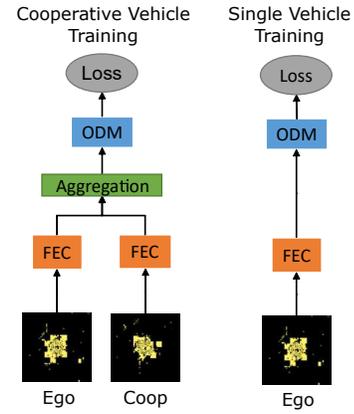}%
\caption{CNN training strategies for the cooperative perception systems}
\label{fig::TrainingStra}%
\end{figure}
\subsubsection{Feature Extraction Component}
Feature Extraction Component (FEC) is a neural network, e.g. CNN, projecting the observations from sensory input state space to intermediary state space. Since the parameters are optimized, FEC provides an optimal representation of the input with respect to the task's objective function.
The compression rate is proportionate to the total number of sub-sampling layers in the FEC network. Although the size of shared feature-maps can be decreased by adding more sub-sampling layers in the FEC network, the performance of object detection will also decline as the down-sampling rate increases. 
Additionally, the last layer of FEC determines the number of feature-map's channels being shared among participants.
Therefore, the network structure of FEC, both by the number of channels and the down-sampling rate, affects the communication requirement of the cooperative system. 
It is worth mentioning that the performance of single-shot detectors can be enhanced by adding shortcut between different CNN layers \cite{Lin_2017_CVPR, He_2016_CVPR}; however, having shortcut between FEC and ODM layers requires transmission of feature-maps created at different layers.
As a result, having shortcut layers between FEC and ODM will significantly increase the size of shared data.
In our realizations of DFS approach presented in this paper, the CNN architecture does not have any shortcut layers.
\begin{figure}[!pt]%
\centering
\subcaptionbox[]{}{%
\includegraphics[width=0.48\textwidth]{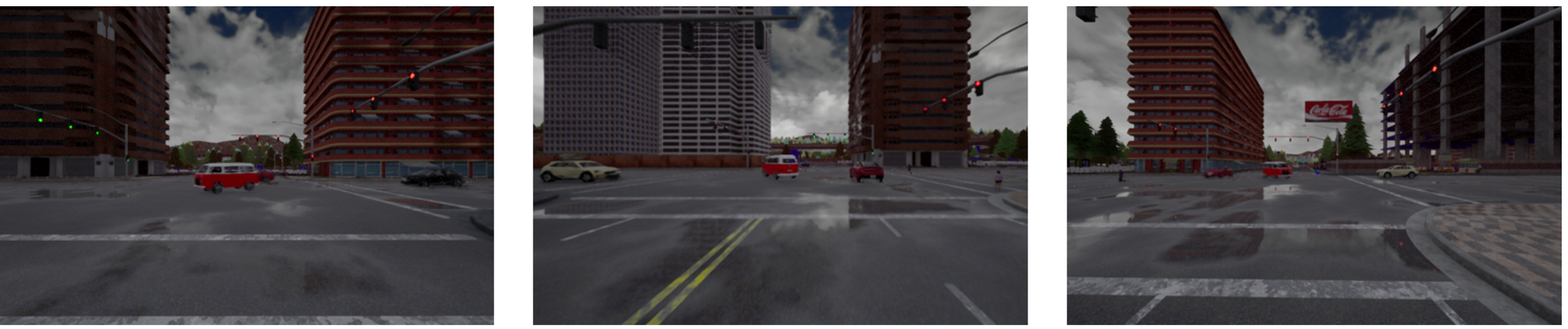}}%
\qquad
\subcaptionbox[]{}{%
\includegraphics[width=0.48\textwidth]{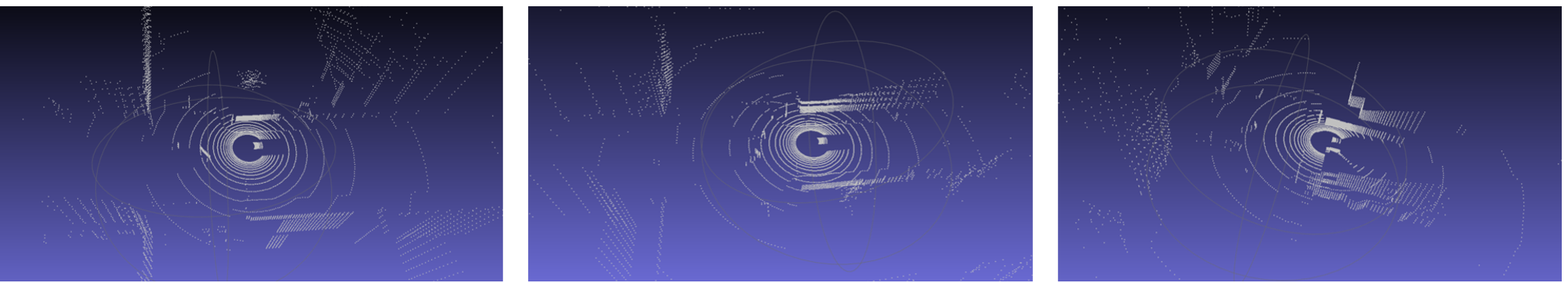}}%
\qquad
% \label{fig:third}%
\subcaptionbox[]{}{%
\includegraphics[width=0.48\textwidth]{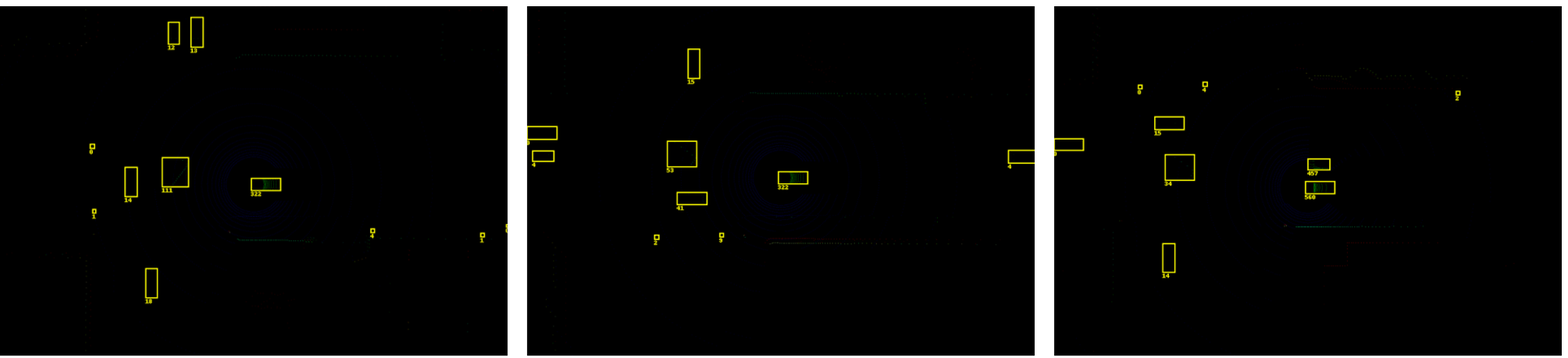}}%
% \label{fig:third}%
\caption{ (a) Illustration of CARLA work-space in an arbitrary scenario (b) Point-cloud generated from LIDAR device of a cooperative vehicle (c) BEV representation of the point-cloud}
\label{fig:foobar}
\end{figure}
\subsubsection{Transmission/Reception Component}
In our previous work, we have shown the advantage of using identical networks in feature-sharing cooperative perception to unify the feature-space among participating observers. Since the feature-space is identical for all participants, the aggregation component can be as simple as an element-wise operation.
As mentioned, the choice of network architecture determines the bandwidth requirement and performance of cooperative systems.
In other words, the size of the shared feature-maps is directly dependant on the FEC network structure.
However, entangling the architecture to bandwidth limitations results in inflexibility of design in adapting to different bandwidth requirements.
In addition, at ego-vehicle, producing feature-maps with a low number of channels will only deteriorate the object detection performance while not having any effect on bandwidth consumption.
To have a flexible design capable of adapting to different bandwidth restrictions, \cite{marvasti2020bandwidthadaptive} proposes to use a set of CNN encoder/decoders, projecting the transmitter vehicle feature maps onto a lower dimension.
\cite{10.1145/3318216.3363300} argues that transmission of a selected subset of channels from the feature-maps can reduce the communication load while maintaining a comparable detection performance.
In summary, the transmission/reception component can address the communication network restrictions and allows the FEC to be designed with more flexibility.

\subsubsection{Aggregation and Alignment}
\label{feturesharing-aggregation}
As it was discussed in section \ref{Section:overallaggregation}, in the DFS methods, a separate procedure is needed to align feature-maps extracted by different participants in addition to translation and rotation transformations.
Although the feature-maps can be aggregated by using translation alignment discussed in \cite{marvasti2020cooperative}, this procedure does not consider the inconsistencies caused by down-sampling the input image.
The Translation MOD-alignment procedure (TMA), introduced in \cite{marvasti2020bandwidthadaptive}, provides a solution to mitigate the misalignment error imposed by down-sampling.
TMA is a decentralized method enabling transmitting vehicles to pre-align the BEV images or input 3D tensors without prior information on the location of the recipient vehicle.
To avoid confusion, the pixels produced in feature-maps are referred to as “fixels”.
The purpose of TMA procedure is to make sure corresponding fixels represent an identical area in the scene. This will ensure that aggregating the fixels will not result in localization error.
Fig.\ref{fig:MA}, originally presented in \cite{marvasti2020bandwidthadaptive}, illustrates how down-sampling leads to localization error.
The illustrated vectors in both feature-maps represent the localization information vector correctly pointing to the target's position. 
However, due to contradictory localization information of the corresponding fixels, the object detection result of aggregated feature-maps can be erroneous.
Assuming that each observer has access to its position, the TMA method solves the aforementioned contradiction by padding the input images or 3D tensors using the following equations:
\begin{equation}
    % \begin{aligned}
(p_l,p_t) \equiv(x_0,y_0) \Mod{K} 
\end{equation}
\begin{equation}
(p_r,p_b)  \equiv (-x_1,-y_1)\Mod{K}
    % \end{aligned}
    \label{eq::mod}
\end{equation}
Where $[x_{0},x_{1}]\times [y_{0},y_{1}]$ denotes the global pixel-wise coordinate range of the input image, and $p_l$, $p_r$, $p_t$, $p_b$, $K$ are left, right, top and bottom padding parameters and down-sampling rate respectively.
Equation (\ref{eq::mod}) shows that TMA is agnostic to the position of receiver-vehicle making it an appropriate alignment procedure in broadcasting feature-maps.

In the DFS methods, aggregation of aligned feature-maps can be done by an arithmetic operation.
\cite{10.1145/3318216.3363300} has proposed to aggregate the features by a max-out function. In \cite{marvasti2020cooperative} an element-wise summation function is used for the aggregation.
However, other aggregation operations can be chosen to fuse the acquired feature-maps.
In this paper, we consider the max-norm function to fuse feature-maps in addition to the element-wise summation and the max-out aggregation functions.  
%we propose two additional aggregation modules and in the next section we compare the performance of using each operator.
% The first operator is averaging the feature-maps and the second is max-norm function.
In the max-norm function, the feature vector with the maximum $l_2$ norm is selected. Assuming $V^{xy}_i$ denotes the corresponding fixels feature-map acquired from vehicle $i$ after alignment at fixel coordinates $x$ and $y$, The norm-max function is defined as:
\begin{equation}
    \hat{V}^{xy} = V^{xy}_k
    % \hat{V} = V_1 I(||V^{xy}_1||>||V^{xy}_2||)+V^{xy}_1 I(||V^{xy}_1||\le||V^{xy}_2||).
\end{equation}
Where k is calculated as:
\begin{equation}
    k = \argmax_i(|V^{xy}_i|)
\end{equation}
$|V_i|$ indicates the $l_2$ norm of the feature vector $V_i$.
In section \ref{section::experiments}, we have compared the performance of the DFS methods using the aforementioned aggregation functions.

\begin{table}[t]
\caption{The CNN Architecture of Cooperative Approaches}
\label{table::archtab}
\begin{center}
% \vspace{-20mm}
\begin{tabular}{ |c| c|| c| c|}

% \multicolumn{4}{c}{Models}& &\\
\hline
% \multicolumn{4}{c}{\textbf{Back-Bone Architecture}}\\
% \hline \hline
\multicolumn{4}{|c|}{\textbf{Raw Data Transmission Layer}} \\ \hline \hline
\multicolumn{4}{|c|}{Input 416x416x3 } \\ \hline
% \multicolumn{4}{|c|}{\textbf{Feature Extraction Component}} \\ \hline
\multicolumn{4}{|c|}{3x3x24 Convolution Batch-Norm Leaky ReLU(0.1)} \\ \hline
\multicolumn{4}{|c|}{Maxpool/2} \\ \hline
\multicolumn{4}{|c|}{3x3x48 Convolution Batch-Norm Leaky ReLU(0.1)}\\ \hline
\multicolumn{4}{|c|}{Maxpool/2} \\ \hline
\multicolumn{4}{|c|}{3x3x64 Convolution Batch-Norm Leaky ReLU(0.1)} \\ \hline
\multicolumn{4}{|c|}{3x3x32 Convolution Batch-Norm Leaky ReLU(0.1)} \\ \hline
\multicolumn{4}{|c|}{3x3x64 Convolution Batch-Norm Leaky ReLU(0.1)} \\ \hline
\multicolumn{4}{|c|}{Maxpool/2} \\ \hline
\multicolumn{4}{|c|}{3x3x128 Convolution Batch-Norm Leaky ReLU(0.1)} \\ \hline
\multicolumn{4}{|c|}{3x3x64 Convolution Batch-Norm Leaky ReLU(0.1)} \\ \hline
\multicolumn{4}{|c|}{3x3x128 Convolution Batch-Norm Leaky ReLU(0.1)} \\ \hline
\multicolumn{4}{|c|}{Maxpool/2} \\ \hline
\multicolumn{4}{|c|}{3x3x128 Convolution Batch-Norm Leaky ReLU(0.1)} \\ \hline
\multicolumn{4}{|c|}{3x3x128 Convolution Batch-Norm Leaky ReLU(0.1)} \\ \hline \hline
\multicolumn{4}{|c|}{\textbf{Feature Sharing Transmission Layer}} \\ \hline \hline
% 1x1xT convolution&1x1xT convolution&1x1xT convolution&1x1xT convolution \\ \hline \hline
% \multicolumn{4}{|c|}{\textbf{Object Detection Component}} \\ \hline \hline
\multicolumn{4}{|c|}{1x1x128 Convolution Batch-Norm Leaky ReLU(0.1)}\\ \hline
\multicolumn{4}{|c|}{3x3x256 Convolution Batch-Norm Leaky ReLU(0.1)}\\ \hline
\multicolumn{4}{|c|}{1x1x512 Convolution Batch-Norm Leaky ReLU(0.1)}\\ \hline
\multicolumn{4}{|c|}{1x1x1024 Convolution Batch-Norm Leaky ReLU(0.1)}\\ \hline
\multicolumn{4}{|c|}{3x3x2048 Convolution Batch-Norm Leaky ReLU(0.1)}\\ \hline
\multicolumn{4}{|c|}{1x1x1024 Convolution Batch-Norm Leaky ReLU(0.1)} \\ \hline
\multicolumn{4}{|c|}{1x1x2048 Convolution Batch-Norm Leaky ReLU(0.1)}\\ \hline
\multicolumn{4}{|c|}{3x3x1024 Convolution Batch-Norm Leaky ReLU(0.1)} \\ \hline
\multicolumn{4}{|c|}{1x1x20 Convolution}\\ \hline
\multicolumn{4}{|c|}{Output 52x52x28}\\ \hline \hline
\multicolumn{4}{|c|}{\textbf{HSM Transmission Layer}} \\ \hline
\end{tabular}
\end{center}
\end{table}
\begin{figure*}[t!]%
  \centering
  \includegraphics[width=1\linewidth,trim={0mm 0mm 0mm 0mm},clip]{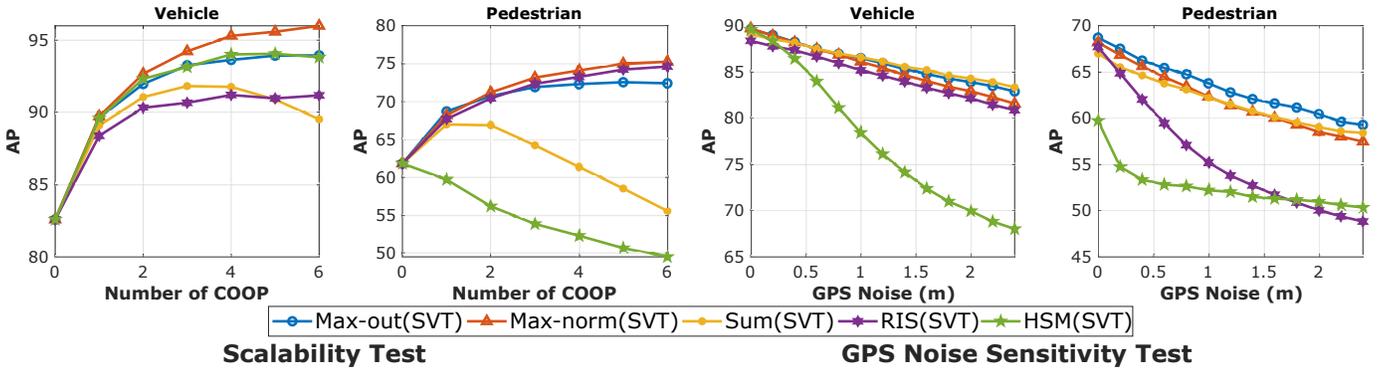}
\caption{The effect of different information aggregation functions with identical network paremeters. The network is trained based on SVT strategy. }
\label{fig::Base}%
\end{figure*}
\begin{figure*}[t!]%
  \centering
  \includegraphics[width=1\linewidth,trim={0mm 0mm 0mm 0mm},clip]{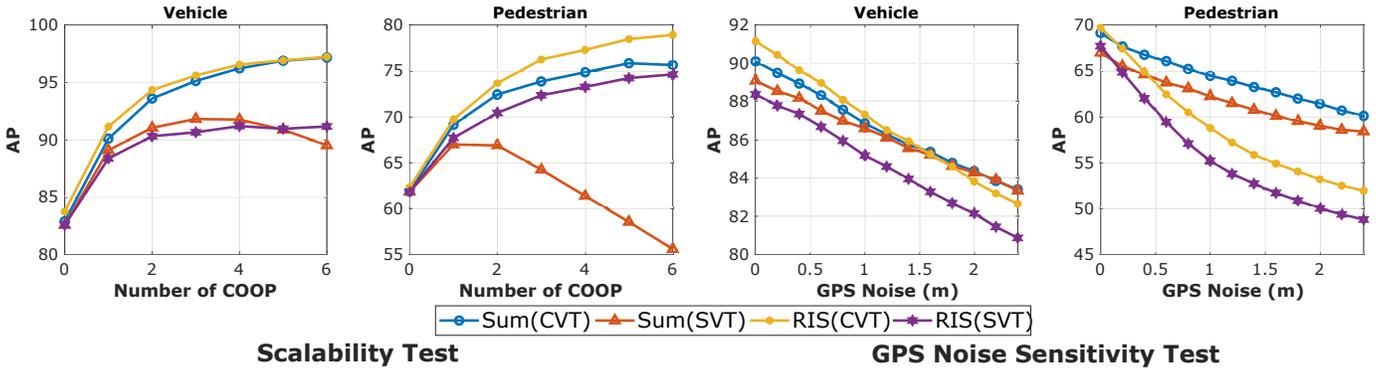}
\caption{The effect of training strategies on the performance of cooperative methods with additive information aggregation functions. }
\label{fig::Train}%
\end{figure*}

\subsubsection{Training Process}
\label{section:trianing}
The training procedures in the DFS methods can significantly affect the performance of cooperative perception systems.
\cite{10.1145/3318216.3363300} showed that a pre-trained single-vehicle object detector can be utilized in the DFS based cooperative perception scheme using the max-out aggregation function. This method of training is referred to as single-vehicle training strategy (SVT) in the rest of the paper.
It is expected that concurrent training of CNNs, where the loss is calculated by aggregation of two cooperative observations, noticeably improves the detection performance. This method was originally proposed in \cite{marvasti2020cooperative} and is referred to as cooperative-vehicle training strategy (CVT). 
In section \ref{section::experiments}, we study the effect of training strategy on the performance of the DFS based cooperative object detection methods with different aggregation functions.

% Furthermore, we propose an additional training strategy for feature sharing approach called cooperative/single vehicle training. In this training strategy, we concurrently train the network via calculating the loss by aggregation of two cooperative observations and add it to the loss calculated from separately feeding each observation to the network.
% In such a manner, the network parameters are trained based on both single-vehicle observations and cooperative vehicle observations.
% Equation \ref{eq::concurrant_training} demonstrates the loss function used for training the networks.
% \begin{equation}
%     L(X_1,X_2;\theta) = L(X_1;\theta)+L(X_2;\theta)+L(X_1\oplus X_2;\theta)
%     \label{eq::concurrant_training}
% \end{equation}
% Where $X_i$, $\theta$ and $\oplus$ indicate the observations, the network parameters and the aggregation function. 
%We refer to this method of training as twin-training.
Fig.\ref{fig::TrainingStra} illustrates the aforementioned training strategies.

% \begin{figure*}[h]
%     \centering
%     \includegraphics[width=1\linewidth,trim={20mm 0mm 30mm 10mm},clip]{Results/Ablation/Scalability/0.7SCALABILITY.eps}
%     \caption{Scalability curve illustrates the performance improvement of different mechanisms by adding cooperative participants. FS-COD with element-wise summation aggregation module is superior in comparison to other approaches}
%     \label{fig:my_label}
% \end{figure*}

% TMA

\section{Experiments and Results}
\label{section::experiments}
To test the performance of cooperative perception methods, we have utilized our cooperative dataset generator tool called Volony\cite{volony2020}.
This tool generates simultaneous observations from the same scene by using CARLA simulator\cite{dosovitskiy2017carla}.
Therefore, we can gather synchronous measurements such as RGBD images and LIDAR point-clouds data.
Additionally, it enables the user to have access to ground-truth information, e.g. bounding-boxes, labels, and vehicles' GPS information.
Moreover, the characteristics of sensors are customizable and various sensors with modifiable range and precision can be deployed on cooperative vehicles.
\begin{figure*}[t!]%
  \centering
  \includegraphics[width=1\linewidth,trim={0mm 0mm 0mm 0mm},clip]{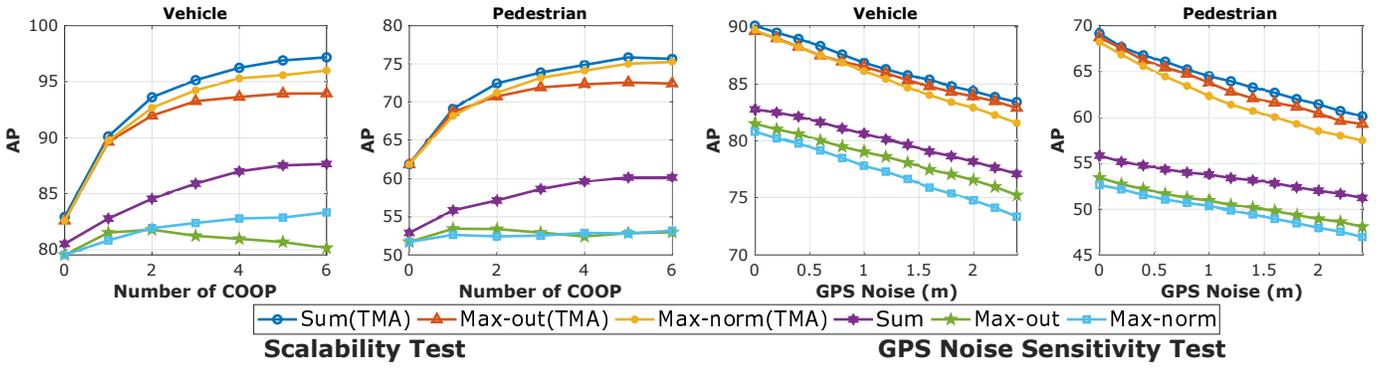}
\caption{The effect of Translation MOD-Alignment on the performance of DFS methods.}
\label{fig::TMARES}%0
\end{figure*}
\begin{figure*}[t!]%
  \centering
  \includegraphics[width=1\linewidth,trim={0mm 0mm 0mm 0mm},clip]{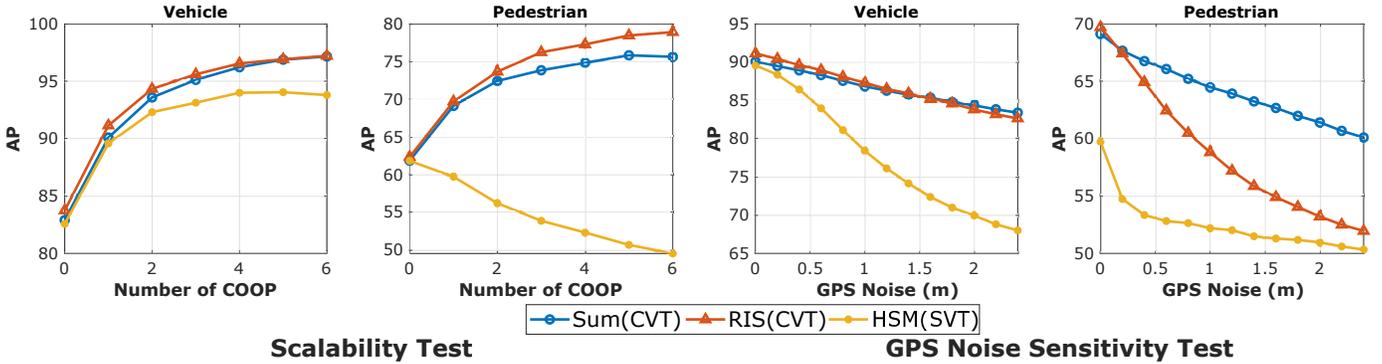}
\caption{Comparison of best performing choices for each information sharing technique. }
\label{fig::CONC}%
\end{figure*}
We have deployed 100 vehicles in an urban area containing roads, buildings, traffic lights, trees, and pedestrians.
Furthermore, 90 vehicles are equipped with LIDAR devices to provide observations.
The urban environment helps to cause complications such as occlusion, yielding a more realistic dataset.
For training and testing, 90000 and 6000 LIDAR observations have been gathered, respectively. These observations are captured at selected 1000 time frames while the simulator was running for 20000 time-steps. This would decrease the correlation between the observed samples at each time index.
In our experiments the task is to detect pedestrians and vehicles in the scene.
In general, we expect to have lower performance in pedestrian detection in comparison to vehicle detection.
By breaking the detection problem to easy and difficult tasks (large vehicles and small pedestrians in size), the distinction among performance of cooperative approaches becomes more evident.
For cooperative training, each LIDAR observation is randomly paired with another observation, gathered by a different vehicle, such that the observers have at least one mutual target in their 40-meter proximity.
To attain fair comparison, the test set is identical for all cooperative perception methods.
% For each training sample, we select a pair of LIDAR equipped vehicles with at least one mutual target in their 40 meter proximity.
As it was discussed, the BEV projection is considered as sensory to input representation component for all cooperative approaches.
The BEV projected image has three channels indicating the density of reflected points at a specific height bin.
The height bins are defined as $[-3m,-1m]$, $[-1m,1m]$ and $[1m,3m]$. The height of a reflected point is calculated relative to the LIDAR device position.
The resolution of the BEV projected image is fixed at $416\times416$ pixels, covering 40 meters at each direction of the observing vehicle.
The Intersection-over-Union (IoU) threshold in the NMS algorithm for merging the hypotheses is set at 40\%.
To have a fair comparison between cooperative approaches, we have chosen identical networks for all methods with the same number of parameters and identical hyper-parameters.
Table \ref{table::archtab} illustrates the CNN architecture used for all methods.
In the RIS method, the raw data is aggregated and fed into the CNN architecture.
In DFS methods, the raw data is separately fed into the first block at different cooperative vehicles. The outputs of this block are shared, aligned, aggregated, and fed into the second block to obtain the detection output.
In the HSM method, the raw data is fed to the entire architecture and the detection output is shared among participants.

In our experiments, average precision is considered as an indicator of cooperative detection performance\cite{padillaCITE2020}. The average precision is calculated based on the IoU threshold of 75\%.
% In order to compare the effectiveness of cooperative approaches, we have chosen average precision as an indicator of detection performance.
% We have assessed the detection performance of these
Furthermore, to provide more insight, we have assessed the sensitivity of detection performances to GPS noise.
Therefore, GPS positioning offsets with fixed magnitude are introduced to participating vehicles' positioning information.
In other words, for each sample, a positioning error vector with uniformly sampled random direction, and the specified magnitude is added to the position of cooperative vehicles.
% At each sample we introduce a vector with the specified magnitude and uniformly sampled random angular direction to the positioning information of cooperative vehicles.
The noise is added only to the coop-vehicles to avoid the scenarios in which ego-vehicle's positioning error cancels out the coop-vehicle positioning error.
We evaluate the noise sensitivity by incrementing positioning error magnitude from 0 to 2.4 meters. The noise sensitivity tests are done by having only one additional cooperative vehicle in the scene.

In addition, we have assessed the effect of cooperative scalability on detection performance.
Our goal is to demonstrate the efficiency of cooperative methods and the corresponding components by increasing the number of participating vehicles equipped with LIDAR.
The test indicates how the performance of cooperative detection is enhanced as more participating entities share their observations.
Additionally, it is important to inspect whether adding more participating vehicles disrupts the performance of the system.

% For the rest of the section, first, we evaluate the performance of feature-sharing techniques with regards to different aspects discussed in the paper and subsequently provide an evaluation of different information sharing approaches in cooperative perception.
% feature-sharing performance compared to other cooperative perception information sharing approaches. 

% \subsection{Results an Discussion}
\begin{figure*}[t!]%
  \centering
  \includegraphics[width=1\linewidth,trim={0mm 0mm 0mm 0mm},clip]{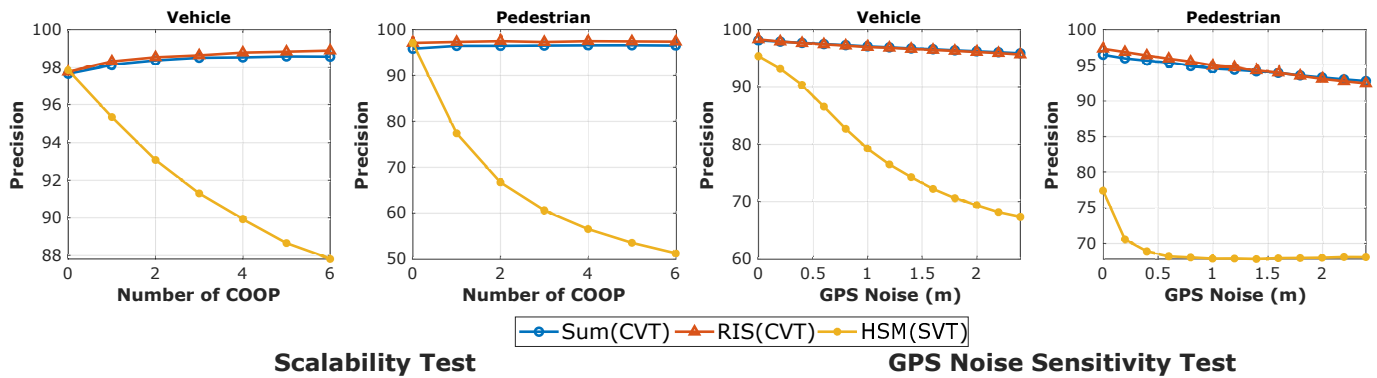}
\caption{Comparison of precision for best performing choices of each information sharing technique. }
\label{fig::CONC_PREC}%
\end{figure*}
\begin{figure*}[t!]%
  \centering
  \includegraphics[width=1\linewidth,trim={0mm 0mm 0mm 0mm},clip]{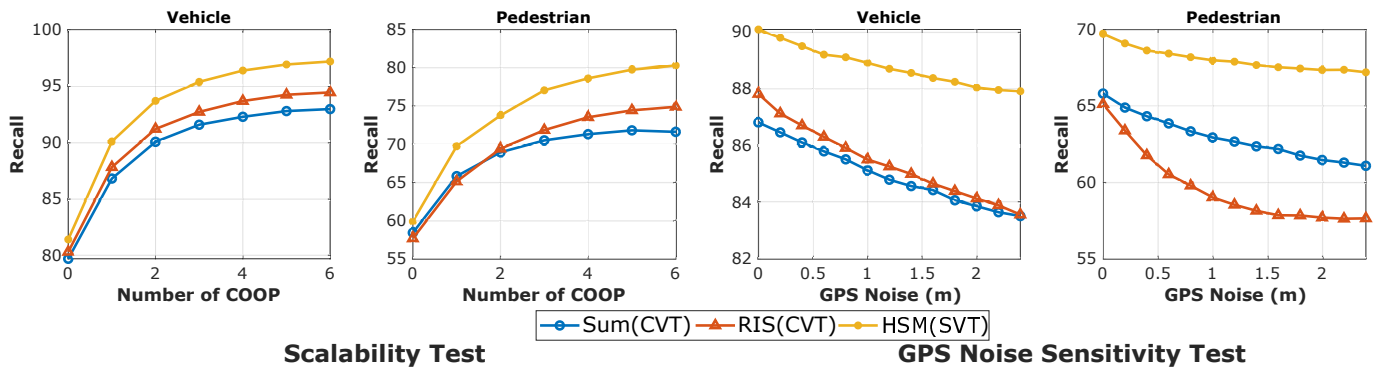}
\caption{Comparison of recall for best performing choices of each information sharing technique. }
\label{fig::CONC_RECALL}%
\end{figure*}
In order to have a thorough evaluation of different cooperative methods, we have tested the DFS methods with different aggregation functions and compared their performance to the RIS and the HSM approaches.
To have a clear perspective on the performance of the methods, we trained a network based on the SVT strategy and used its parameters for all cooperative approaches.
The validation results of each approach are presented in Fig.\ref{fig::Base}. 
The results confirm that DFS methods are relatively more robust to localization error cause by GPS noise.
The robustness of DFS methods to GPS noise is more evident in the detection of pedestrians.
The scalability test in Fig.\ref{fig::Base}, illustrates that the DFS method with the max-norm aggregation module has more performance gain as the number of cooperative vehicles increases.
Additionally, this result shows that the performance of the DFS method with element-wise summation is severely disrupted as the number of cooperative vehicles increases.
Since the network is not trained based on addition, the feature-maps and subsequent CNN layers of the network are not adapted to such a scheme.
Hence, the demonstrated results can be expected for the cooperative method where the aggregation has an additive nature.
Similarly, in the RIS method, the integration of point-clouds acquired from cooperative vehicles is equivalent to the summation of BEV maps.
Therefore, we suspect that the RIS method can suffer from the same issue.
As a solution, we have utilized the CVT strategy discussed in section \ref{section:trianing} to remedy the performance degradation.

Fig.\ref{fig::Train} demonstrates the effect of training strategy on the RIS and the DFS with element-wise summation aggregation methods. The result confirms that by training the networks in a cooperative manner, not only the scalability problem is solved but also the cooperative object detectors have a boost in their performance.
In our experiments, we used one concurrent cooperative observation in the CVT training phase.
These results suggest that including concurrent observations gathered by more cooperative vehicles in the training phase would further improve the generalization of the networks. 

In order to study the effect of the alignment procedure, we have removed the the TMA component and replaced it with translation and rotation transformations.
Fig.\ref{fig::TMARES} depicts the performance of the DFS methods with and without the TMA component.
The results indicate that utilizing the TMA component would significantly improve the performance of the networks with respect to scalability.
Additionally, omitting this component would drastically affect the performance of DFS methods with the max-out and max-norm aggregation functions.
Since methods utilizing maximum operator have selective nature, it is more probable to select the fixle or feature-channel of cooperative vehicles as the number of participating vehicles increases.
Since feature-maps of cooperative vehicles have the misalignment error, adding more participants cooperating with each other does not necessarily improve performance.
On the other hand, we speculate that the misalignment error will be neutralized by using element-wise summation aggregation function as the number of cooperative vehicles increases.

To summarize, Fig.\ref{fig::CONC} compares the best performing networks to provide a conclusive insight on cooperative approaches.
As depicted, RIS has the best performance in the scalability test.
However, in the scalability test, the performance of the DFS method with the TMA component and element-wise aggregation function is slightly lower than the performance of the RIS method. 
Furthermore, the DFS method with max-norm aggregation function shows promising results considering that it does not require to be trained cooperatively.
Additionally, feature-sharing approaches are significantly more robust to GPS noise in the case of pedestrian detection.
The performance of HSM approach is relatively poor in both scalability and GPS noise sensitivity tests.
In pedestrian detection, HSM performance drops by increasing the number of cooperative vehicles.
To investigate the reason for performance drop we have presented the precision and recall measures of selected cooperative approaches in Fig.\ref{fig::CONC_PREC} and Fig.\ref{fig::CONC_RECALL} respectively.
In Fig.\ref{fig::CONC_PREC}, we can observe that the precision performance of HSM drops as GPS noise magnitude increases. The same trend is observed in scalability test.
Also, the precision performance decrease is more severe in the case of pedestrian detection.

Here, we discuss how current NMS designs used for merging the hypotheses in single-shot networks lead to the precision drop effect in the HSM approach.
Single-shot object detectors have a limited number of hypotheses per output cell.
In our case, the number of hypotheses is set to 4.
Therefore, in the RIS and the DFS approaches, where the aggregation of data happens before proposing the hypotheses, the total number of hypotheses used for the calculation of average precision is not dependent on the number of cooperative vehicles.
On the contrary, in the HSM approach, where cooperative vehicles share their independently derived hypotheses, the total number of hypotheses is dependent on the number of cooperative vehicles.
Therefore, as the number of participants increases, a drastic drop in precision performance can be caused by two factors in the HSM approach.

First, if the shared hypotheses are not merged properly the number of false-positive cases increases as more cooperative participants share their hypotheses. In other words, if the hypotheses that are pointing to the same target in the scene are not merged correctly all except one are going to be considered as false positives.
In the case of pedestrian detection, the IoU threshold of 40\% for NMS means that the localization error should be approximately within the vicinity of 0.2 meters. Hence, the NMS method is more prone to errors in the merging of pedestrian detection hypotheses.
This argument is true with regard to increasing the GPS noise magnitude.
Second, in HSM, due to the nature of the NMS algorithm, if cooperative vehicles share false-positive hypotheses, NMS at best merges such false hypotheses.
Therefore, by increasing the number of cooperative vehicles, false-positive hypotheses of participating vehicles are accumulated.
This issue does not occur in other approaches as observed in detection precision curves presented in Fig.\ref{fig::CONC_PREC}.
This is due to having 4 anchors per cell in the design of single-shot networks, i.e. having a limited number of proposed hypotheses independent from the number of participating vehicles, and elimination of false positives during the fusion process.
We can observe that the precision increases in the DFS with element-wise summation function and RIS method.
Additionally, it is worth noting that the HSM approach relatively has higher recall performance as presented in Fig.\ref{fig::CONC_RECALL}.
The higher recall performance can be the consequence of not having hypotheses limitation as opposed to the DFS and the RIS methods.

In conclusion, we should note that all cooperative methods used in this paper utilize identical NMS functions with identical hyper-parameters.
Therefore improving NMS will lead to improving all methods.
However, the precision results indicate the importance of NMS and further elaborates possible weaknesses of the current design of NMS.
Hence, it is vital to revisit NMS functionality design to properly address the issues arising in cooperative approaches.

\section{Concluding Remarks}
\label{Section:Concluding Remarks and Future work}
We have evaluated the performance of three categories of cooperative perception approaches by analyzing different cooperative object detection designs in this paper. Our results confirm that DFS methods outperform HSM in terms of average precision. In addition, The detection performance boost caused by increasing the number of cooperative vehicle in arbitrary scenes in the DFS method with element-wise summation aggregation function is almost similar to the RIS method, while DFS method is significantly less sensitive to localization error (GPS noise) compared to the RIS method. Moreover, by enabling the cooperative participants to effectively compress and encode relevant information, the DFS methods have significantly lower communication requirement compared to RIS method while maintaining the object detection performance at a desirable level.

The promising improvement resulting from utilizing the novel concept of Deep Feature Sharing motivates us to further investigate its application in multi-modal sensing approaches to effectively share and integrate volumetric and monocular sensory data. Utilizing such inference methods in a cooperative context using DFS approach requires developing new feature extraction, alignment and aggregation designs.

% \section{Discussion}
% \section{Conclusion}
\bibliographystyle{IEEEtran}\small
\bibliography{main}
\end{document}